\documentclass[10pt,twocolumn,letterpaper]{article}

\usepackage{cvpr}
\usepackage{times}
\usepackage{epsfig}
\usepackage{graphicx}
\usepackage{amsmath}
\usepackage{amssymb}

\usepackage{graphicx}
\usepackage{makecell}
\usepackage{multirow}
\usepackage{subcaption}
\usepackage{authblk}
\usepackage{epigraph}

\usepackage[breaklinks=true,bookmarks=false]{hyperref}

\cvprfinalcopy 

\def\cvprPaperID{****} 

\ifcvprfinal\fi
\begin{document}


\title{TransGaGa: Geometry-Aware Unsupervised Image-to-Image Translation\vspace{-0.6cm}}

\author{Wayne Wu$^1$ Kaidi Cao$^2$ Cheng Li$^1$ Chen Qian$^{1}$ Chen Change Loy$^{3}$ \\
$^1$SenseTime Research \hspace{10pt} $^2$Stanford University \\
$^3$Nanyang Technological University \\
{\tt\small \{wuwenyan, chengli, qianchen\}@sensetime.com}\hspace{1cm}
{\tt\small kaidicao@cs.stanford.edu}\hspace{1cm}
{\tt\small ccloy@ntu.edu.sg}
}

\twocolumn[{%
\renewcommand\twocolumn[1][]{#1}%
\vspace{-3em}
\maketitle
\vspace{-5em}
\begin{center}
    \centering
    \includegraphics[width=0.95\linewidth]{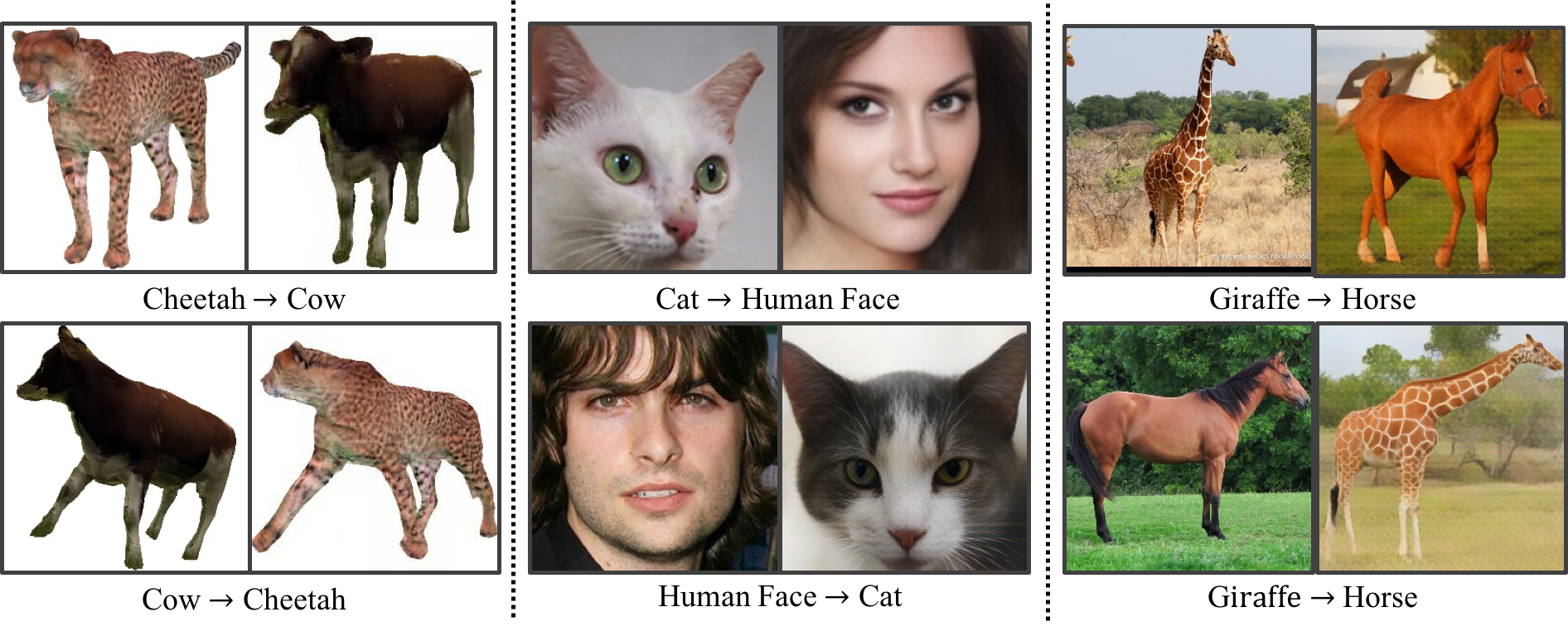}
    \vspace{-0.7em}
    \captionof{figure}{\small{We propose a geometry-aware framework for \textit{unsupervised} image-to-image translation, which is robust to arbitrary shape variations between domains. We show the results of both near-rigid and non-rigid objects. \textit{(left)} Cow and cheetah rendered from CAD models. \textit{(center)} Cat and human face from in-the-wild datasets. \textit{(right)} Horse and giraffe from Flickr.}
    }
\end{center}%
}]

\maketitle
\thispagestyle{empty}

\begin{abstract}
Unsupervised image-to-image translation aims at learning a mapping between two visual domains. However, learning a translation across large geometry variations always ends up with failure. In this work, we present a novel disentangle-and-translate framework to tackle the complex objects image-to-image translation task. Instead of learning the mapping on the image space directly, we disentangle image space into a Cartesian product of the appearance and the geometry latent spaces. Specifically, we first introduce a geometry prior loss and a conditional VAE loss to encourage the network to learn independent but complementary representations. The translation is then built on appearance and geometry space separately. Extensive experiments demonstrate the superior performance of our method to other state-of-the-art approaches, especially in the challenging near-rigid and non-rigid objects translation tasks. In addition, by taking different exemplars as the appearance references, our method also supports multimodal translation. Project page: \color{red}{\url{https://wywu.github.io/projects/TGaGa/TGaGa.html}}
\end{abstract}

\section{Introduction}

\epigraph{I will be your mirror. Reflect what you are, in case you don't know. I will be the wind, the rain and the sunset. The light on your door to show that you are home.}{\textit{Lou Reed}}
\vspace{-0.3cm}
\textit{Unsupervised} image-to-image translation aims at learning tahe translation between two different image domains without any pairwise supervision. 
The notion of image translation has been widely applied in colorization~\cite{DBLP:conf/eccv/ZhangIE16}, super-resolution~\cite{DBLP:conf/cvpr/LedigTHCCAATTWS17,wang2018esrgan} and style transfer~\cite{DBLP:conf/nips/GatysEB15}.

Early works demonstrated the effectiveness of deep neural network in transferring local textures, demonstrating successful cases on seasonal scene shifting~\cite{CycleGAN2017,mingyu2017unsupervised} and painting style transfer~\cite{hsinying2018diverse}. However, researchers soon realized its limitation on the more complicated cases, \ie, translation between two domains with large geometry variations~\cite{CycleGAN2017,gokaslan2018improving}. 
To handle more complex cases, one has to establish the translation on the higher semantic level. For example, based on the understanding of the components of neck, body and leg of a horse, we may imagine a giraffe with the same posture. However, one can hardly implement this translation by replacing the local texture due to the large geometry variations between the two domains.

Performing a translation on the higher semantic level is non-trivial. Geometry information plays a critical role here but, often, there is a significant geometry gap between two image domains, \eg, cat to human-face and horse to giraffe. Although containing the same corresponding components with the similar semantic meaning between the two domains, their spatial distributions are rather different.

In this paper, we propose a novel geometry-aware framework for unsupervised image-to-image translation. Instead of directly translating on the image space, we first map the image into the Cartesian product of geometry and appearance spaces and then perform the translation in each latent space. 
To encourage the disentanglement of two spaces, we propose an unsupervised conditional variational AutoEncoder framework, in which a Kullback-Leibler (KL) divergence loss and skip-connection design are introduced to encourage the network to learn a complementary representation of geometry and appearance. Then we build the translation between two domains based on their bottleneck representation. Extensive experiments show the effectiveness of our framework in establishing translation between objects both on synthesis and real-world datasets. Our method achieves superior performance to state-of-the-art methods in both qualitative and quantitative experiments.

We summarize the contributions of this work as follows:
1) We propose a novel framework for unsupervised image-to-image translation. Instead of directly translating on the image space, we build the mapping between two domains on their disentangled latent appearance-geometry space. Our framework extends the ability of CycleGAN on more complicated objects like animals.
2) Fine-disentangled latent space naturally endows our model with the ability of diverse and exemplar-guided generation, which is a challenging and ill-posed multimodal problem in unsupervised image-to-image translation.

\section{Related Work}

\noindent
\textbf{Image-to-Image Translation.} The goal of image-to-image translation is to learn a mapping from a source image domain to a target image domain. Pix2Pix~\cite{isola2017image} proposes a unified framework for image-to-image translation first time based on conditional GANs. Several works~\cite{wang2018pix2pixHD,wang2018vid2vid} extend it to deal with high-resolution or video synthesis. Although appealing results have been shown, these methods need paired data for training. For unsupervised image-to-image translation with unpaired training data, CycleGAN~\cite{CycleGAN2017}, DiscoGAN~\cite{taeksoo2017learning}, DualGAN~\cite{zili2017dualgan} and UNIT~\cite{mingyu2017unsupervised} are proposed based on the idea of cycle-consistency. GANimorph~\cite{gokaslan2018improving} introduce a discriminator with dilated convolutions to get a more context-aware generator. However, without paired training data, the translation problem is inherently ill-posed because of the infinite existing mappings between two domains.  Recent studies have attempted to solve this problem for multimodal generations. CIIT~\cite{Lin_2018_CVPR}, MUNIT~\cite{xun2018multimodal}, DRIT~\cite{hsinying2018diverse} and EG-UNIT~\cite{liqian2018exemplar} decompose the latent space of images into a domain-invariant content space and a domain-specific style space to get diverse outputs. However, once the cross-domain structure variation becoming large, the assumption of domain-invariant content space is violated. Even though it is intuitive to share the latent space of content across domains in style transfer tasks, it is hard to embed the complex geometry cues of different domains with one shared distribution. The performance of all existing methods degrades dramatically in the translation with large cross-domain geometry variations.

\noindent
\textbf{Structural Representation Learning.} To model visual content, several unsupervised techniques have been proposed including VAE~\cite{Kingma2014}, GANs~\cite{ian2014generative} and ARNs~\cite{oord2016pixel,NIPS2016_6527}. Recently, many literature focus on unsupervised landmark discovery~\cite{james2017unsupervised,DBLP:conf/nips/ThewlisBV17,Zhang_2018_CVPR,jakab2018conditional,chen2019self} for structural representation learning. Since landmark is an explicit representation for the structure of objects, it can better capture the intrinsic shape of object than other representations. Inspired by the recent development of unsupervised landmark discovery, a heatmap-stack of landmarks are learned in this work for explicit structure representation.

\noindent
\textbf{Disentanglement of Representation.} Disentanglement is important for the control over structure and appearance. There exist a  number of studies on face and person image generation~\cite{Balakrishnan_2018_CVPR,esser2018variational,ma2018disentangled,Wang_2018_CVPR}. Although enjoying the advantage of well pose-guided synthesis, these methods require pre-defined annotations for supervised learning. Several works for unsupervised disentanglement have been proposed, \eg, InfoGAN~\cite{chen2016infogan} and $\beta$-VAE~\cite{hadditioniggins2017beta}. However, these methods suffer from the lack of interpretability, and the meaning of each learned factor is uncontrollable. Instead, our method is able to obtain a controllable disentanglement of structure and appearance in a completely unsupervised manner.

\begin{figure}[t]
\centering
\includegraphics[width=0.9\linewidth]{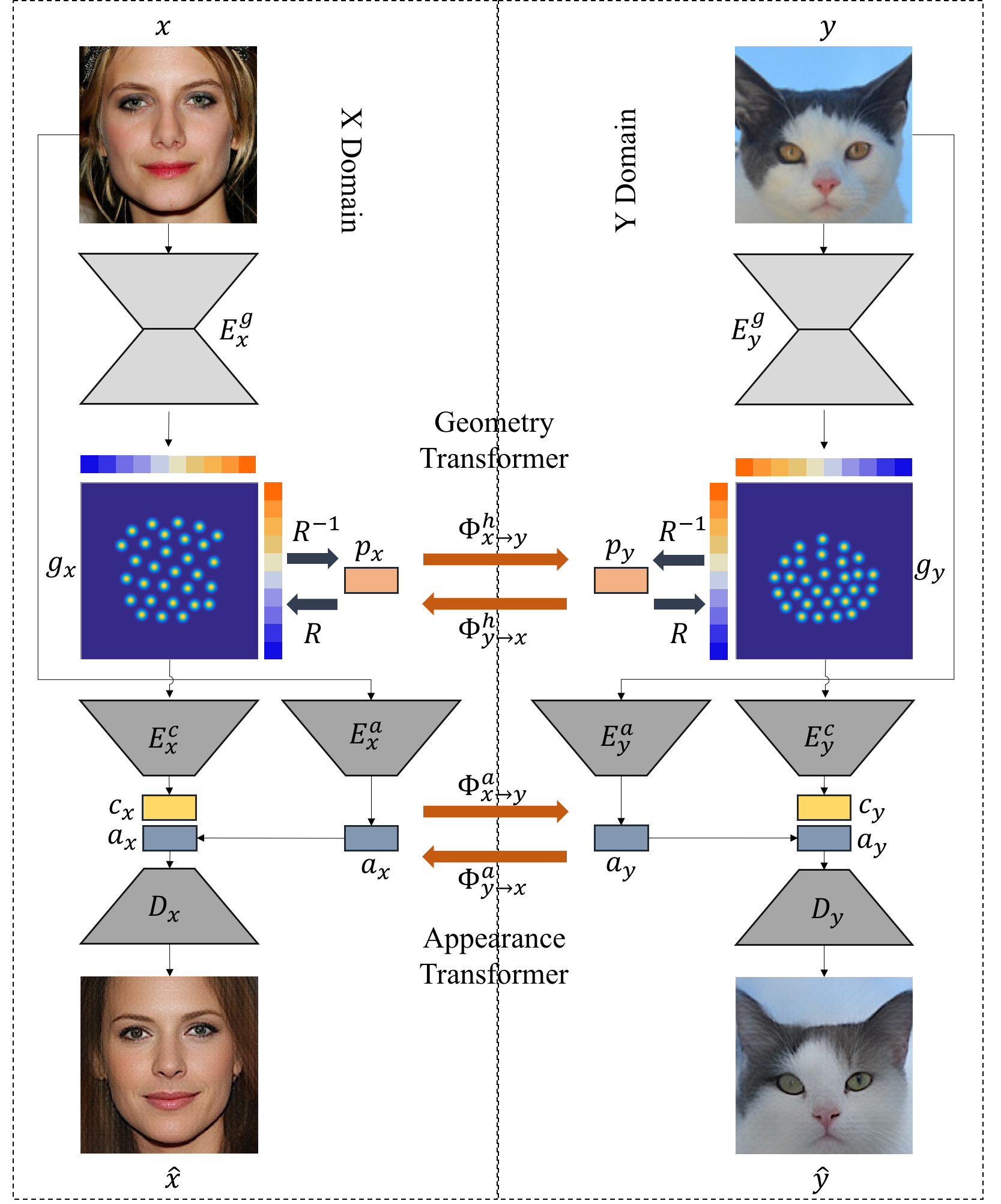}
\caption{\small{\textbf{Architecture.} Our framework consists of four main components: two auto-encoders (X/Y domain) and two transformers (geometry/appearance). \textit{Auto-Encoder}: Taking X domain for example. For the input $x$, we use an encoder $E^g_x$ to obtain the geometry representation $g_x$, which is a 30-channel point-heatmap with the same resolution as $x$. We project all channels of $g_x$ together for visualisation. Then, $g_x$ is embedded again to get the geometry code $c_x$. At the same time, $x$ is also embedded by appearance encoder $E^a_x$ to get the appearance code $a_x$. Finally, $a_x$ and $c_x$ are concatenated together to generate $\hat{x}$ with $D_x$. \textit{Transformer}: For cross-domain translation, geometry ($g_x\leftrightarrow g_y$) and appearance ($a_x\leftrightarrow a_y$) transformation are performed separately.}}
\vspace{-15pt}
\label{fig:fig_architecture}
\end{figure}

\section{Methodology}

Given two image domains $X$ and $Y$. The goal of our work is to learn a pair of mapping $\Phi_{X  \rightarrow Y}$ and $\Phi_{Y  \rightarrow X}$ that could transfer an input $x \in X$ to a sample $y=\Phi_{X  \rightarrow Y}(x)$, $y \in Y$, and vice versa. This problem formulation is a typical unpaired cross-domain image translation task, where the biggest challenge lies in tasks that require geometric changes \cite{CycleGAN2017,gokaslan2018improving}. Most existing frameworks try to parameterize these pairs of mapping through two neural networks, \eg, ResNet~\cite{DBLP:conf/cvpr/HeZRS16} or HourGlass~\cite{alejandro2016stack}.], of which the optimization is hard under complicated scenarios. 
In this study, we assume each domain can be disentangled into a Cartesian of structure space $G_\cdot$ and appearance space $A_\cdot$. Then on each space, we build a transition between the two domains, \ie, \textit{geometry transformer} $\Phi^g_{X  \rightarrow Y}$ and $\Phi^g_{Y  \rightarrow X}$ for geometry space and \textit{appearance transformer} $\Phi^a_{X  \rightarrow Y}$ and $\Phi^a_{Y  \rightarrow X}$ for appearance space. Figure~\ref{fig:fig_architecture} illustrates the framework of our proposed approach.

\subsection{Learning Disentangled Structure and Style \\ Encoders}

Unlike previous works that employ an encoder-decoder structure aiming at encoding all the information using one convolutional network~\cite{CycleGAN2017,DBLP:conf/nips/ZhuZPDEWS17}, our approach tries to encode geometry structure and the appearance style separately. To achieve this, we apply a conditional variational autoencoder in each domain. The conditional VAE system consists of an unsupervised geometry estimator $E^g_\cdot(;\pi)$, a geometry encoder $E^c_\cdot(;\theta)$ which embeds the heatmap structure into the latent space $C_\cdot$, an appearance encoder $E^a_\cdot( ; \phi)$ which embeds the appearance information into the latent space $A_\cdot$, and a decoder $D_\cdot(;\omega): C_\cdot \times A_\cdot \rightarrow X/Y$, which maps the latent space back to the image space. To disentangle two representations in an unsupervised manner, we formulate our loss as the combination of a conditional VAE loss and a prior loss for geometry estimation, which is
\begin{equation}
  \mathcal{L}_{\mathrm{disentangle}} =  \mathcal{L}_{\mathrm{CVAE}} + \mathcal{L}_{\mathrm{prior}}.
\end{equation}
Inspired by previous literature~\cite{Kingma2014,sohn2015learning,esser2018variational}, we implement the conditional VAE loss as:
\begin{equation}
\begin{aligned}
  \mathcal{L}_{\mathrm{CVAE}}(\pi,\theta, \phi, \omega) = - KL(q_\phi(c|x,g) || p(a|x) ) \\
  + \| x - D(E^c (E^g(x)), E^a(x) ) \| ,
\label{eq:lossprior}
\end{aligned}
\end{equation}
where the first term is the KL-divergence loss between two parametric Gaussian distributions and the second term is a reconstruction loss. Here we replace it with the perceptual loss of a VGG-16~\cite{DBLP:journals/corr/SimonyanZ14a} network. In the supervised manner, $\mathcal{L}_{\mathrm{CVAE}}$ can facilitate the learning of a complementary representation of geometry and appearance as described in~\cite{esser2018variational}. However, in our unsupervised scenario, one cannot guarantee any branch of encoders to learn the geometry information without the supervision of geometry map $g_\cdot$. Next we will introduce our prior loss to constrain the geometry estimator.

\subsection{Prior Loss for Geometry Estimator}

Contrary to existing literature that use a content encoder to embed all of the detailed contents~\cite{mingyu2017unsupervised,hsinying2018diverse}, our geometry estimator $E^g_\cdot$ tries to distil pure geometry structure information as a stack of landmark heatmap. To achieve this, we rely on prior knowledge of how object landmarks should distribute to constrain the learning of our structure estimator $E^g_x$ and $E^g_y$ as described in \cite{Zhang_2018_CVPR,jakab2018conditional}. These previous work has shown that it is possible when given appropriate prior losses and learning architecture.

We now introduce the set of prior losses we used: 
\begin{align}
    \mathcal{L}{\mathrm{prior}} = \sum_{i \neq j} \exp (-\frac{ ||g^i - g^{j} ||^2}{2 \sigma ^ 2}) + \text{Var}(g)
\label{eq:lossprior}
\end{align}
The first term is a \textit{Separation Loss}. Similar to the difficulty described in \cite{Zhang_2018_CVPR}, we find that training the structure branch with general random initialization tend to locate all structural points around the mean location at the center of the image. This could lead to a local minimum from which optimizer might not escape. As such, we introduce the separation loss to encourage each heatmap to sufficiently cover the object of interest. This is achieved by the first part in Eq. \ref{eq:lossprior}, where we encourage each pair of $i^{th}$ and $j^{th}$ heatmaps to share different activations. $\sigma$ can be regarded as a normalization factor here. 
The second term is a \textit{Concentration Loss}, which we introduce to encourage the variance of activations $g$ to be small so that it could concentrate at a single location. This corresponds to the second term in Eq. \ref{eq:lossprior}.

\begin{figure}[t]
\centering
\includegraphics[width=0.85\linewidth]{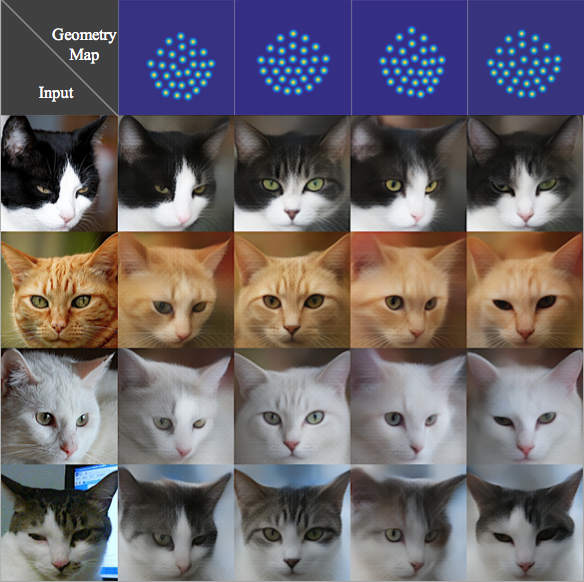}
\caption{\small{\textbf{Disentangled representation.} The top-row shows the corresponding geometry heatmaps of the faces in the left-most column. We illustrate the explicitly disentangled latent space with a grid of structure$\&$appearance swapping results. In each column, the shape of the generated images is shown to be consistent with the geometry heatmaps. In each row, the appearance of the generated images are shown to be consistent with the left-most ones.}}
\vspace{-0.4cm}
\label{fig:fig_disentangle}
\end{figure}

The geometry prior, which is an explicit presentation of object shape, is important for a fine disentanglement of appearance and geometry. As shown in Fig.~\ref{fig:fig_disentangle}, with the geometry maps as the conditional input, our method can generate different shapes of face, which are consistent with geometry maps while maintaining the appearance of one specific input. It indicates that by estimating the pure geometry cues of objects, our method can disentangle geometry and appearance within a domain in a completely unsupervised manner.

\subsection{Appearance Transformer}

With the disentangled appearance geometry space, we can discompose the image translation into two separate problems. In this section, we first consider the transformation $\Phi^a$ on the appearance latent space $A_X$ and $A_Y$. 
One may address this latent to latent transformation problem as a CycleGAN~\cite{CycleGAN2017}, with the cycle consistency loss and the adversarial loss. However, this does not guarantee $g_x$ and mapped appearance transformer $\Phi_{X \rightarrow Y}(g_x)$ associated with two images to have a visual relationship. Since these two constraints can only lead to a translation between two distributions, which is arbitrary and multimodal. To this end, we introduce a cross-domain appearance consistency loss to constrain the appearance transformer:
\begin{equation}
  \mathcal{L}^a_{\mathrm{con}}  = \| \zeta( x )  - \zeta(D_y\left(\Phi^g_{x \rightarrow y} \cdot E^g_x (x ) , \Phi^a_{x \rightarrow y} \cdot E^a_x (x ) \right) ) \|,
\end{equation}
where $\zeta$ is the \textit{Gram matrix}~\cite{DBLP:conf/nips/GatysEB15,DBLP:conf/eccv/JohnsonAF16} calculated with a pre-trained VGG-16~\cite{DBLP:journals/corr/SimonyanZ14a} network, $\Phi^g_{x \rightarrow y} \cdot E^g_x (x )$ is the geometry code transformed from X to Y, $\Phi^a_{x \rightarrow y} \cdot E^a_x (x )$ is the appearance code transformed from X to Y, and $D_y(,)$ refers to decoder of Y domain. This loss ensures the image associated with  $g_x$ and translated appearance  $\Phi_{X \rightarrow Y}(g_x)$  to have a similar appearance. In our experiment, it is observed that CycleGAN without appearance constraint can also converge, but it yields different results in each time of training with the same settings. The appearance consistency constraint stables the training and provide a more explainable results.

\noindent
\textbf{Single and Multimodal Transition:} In our framework, the transform function is learned both in appearance and geometry latent spaces. For single-modal translation, the appearance transform $\Phi^a$ is constrained to guarantee transformed samples to have an associated appearance on the image domain. However, as aforementioned, a complex transformation problem is always multimodal. In our method, by replacing the transformed appearance representation by any feasible vector in the target appearance space $A$, we can achieve the results for multimodal generation. For example, with only the geometry transform $\Phi^g$, by taking different human faces as a reference, we can obtain different results by just one cat face input. The multimodal ability is brought by the fine-disentangled representation within the domain. Qualitative results are shown in Sec.~\ref{sec:disentagleExperiment}.

\begin{figure*}[t!]
\centering
\includegraphics[width=1.0\linewidth]{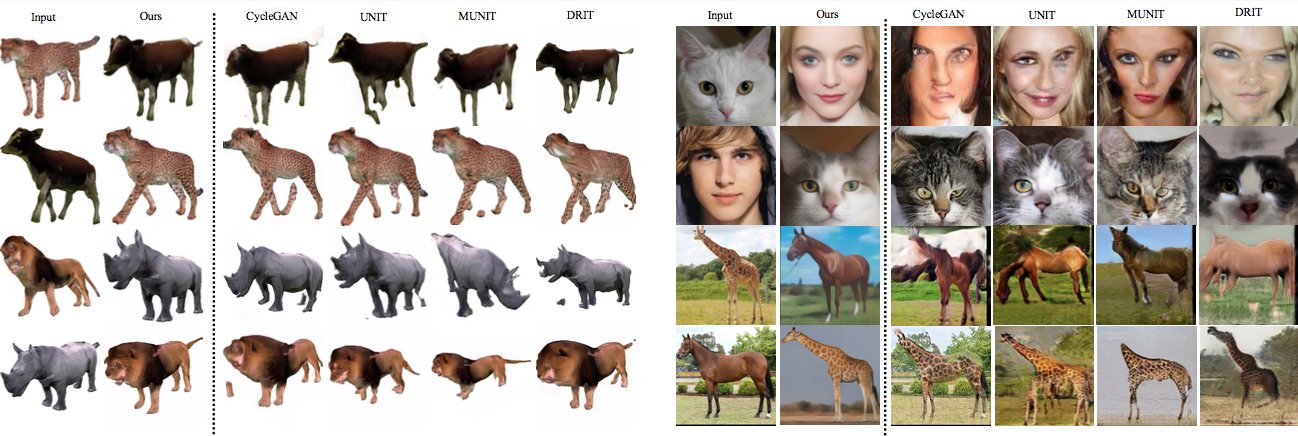}
\vspace{-0.45cm}
\caption{\small{\textbf{Comparison in geometry-preserving.} Results on (a)  synthesis datasets (cow$\leftrightarrow$cheetah and lion$\leftrightarrow$rhino) (b) real-world datasets (cat$\leftrightarrow$human face and giraffe$\leftrightarrow$horse). From left to right: input, ours, CycleGAN~\cite{CycleGAN2017}, UNIT~\cite{mingyu2017unsupervised}, MUNIT~\cite{xun2018multimodal} and DRIT~\cite{hsinying2018diverse}.}}
\vspace{-0.45cm}
\label{fig:fig_compare_structure}
\end{figure*}

\begin{figure*}[t!]
\centering
\includegraphics[width=1.0\linewidth]{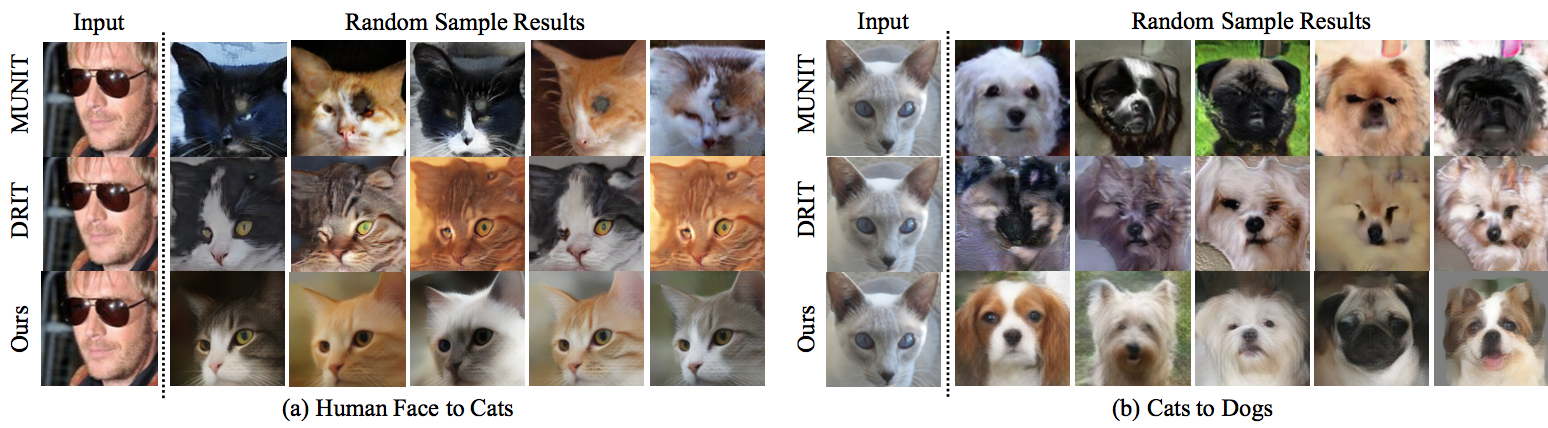}
\vspace{-0.45cm}
\caption{\small{\textbf{Comparison in multi-modal generation.} Results on (a)  human$\rightarrow$cat face (b) cat$\rightarrow$dog face. From top to bottom: MUNIT\cite{xun2018multimodal}, DRIT~\cite{hsinying2018diverse} and ours (zoom in for more details).}}
\vspace{-0.5cm}
\label{fig:fig_compare_multimodal}
\end{figure*}

\subsection{Geometry Transformer}

We found it hard to learn a transfer between unsupervised learned geometry heatmaps directly since CNNs are usually not well-suited at capturing geometry information. Instead, we extract the coordinates information of each landmark from the heatmaps directly with the differentiable re-normalisation operator~\cite{jakab2018conditional} $R$. Thus, the \textit{de facto} geometry transformation is performed in the landmark coordinate space.

Specifically, for each landmark's heatmap, we compute a weighted average of coordinates over all activations across each heatmap. Although the dimensionality of landmarks with 2D coordinates is lower than the image representation, we still use the PCA to reduce dimensions of the landmark representation. The reason behind it is that we observe the result is more sensitive to small errors in geometries than in image pixel values, since slight errors of coordinates may cause severe artifacts (\eg foldover and zigzag contour). It indicates that geometry translation is sometimes harder than image translation.

It is noteworthy that we have tried three kinds of representations for Geometry Transformer (\ie, geometry heatmaps, landmark coordinates and PCA embedding of coordinates). All of the three representations can be used for training in our experiments. PCA embedding of coordinates works best in terms of \textit{stability} and \textit{convergence} of model training while other representations sometimes fail in some specific tasks.
PCA constrains the geometry structure in the output. It constructs an embedding space for geometry shapes, where each principal component represents a reasonable dimension. Therefore, any sample in the embedding space will maintain the basic object structure, which reduces the risk of mode collapse.

To incorporate the PCA landmark representation with GAN, we replace all Conv-ReLU blocks with FC-ReLU blocks in both generators and discriminators. Though we incorporate a similar transformer structure as in CariGANs~\cite{cao2018cari}, our work differs in that unlike CariGANs that uses landmarks' PCA embeddings directly as the source and target domain defined in CycleGAN, we train the corresponding cycle on image pixel level as discussed in Sec.~\ref{other_constraints}, which is more direct and powerful for pose-preserving generation task.

\label{other_constraints}
\subsection{Other Constraints}
Other than the proposed geometry prior loss and style consistency loss, we also leverage cycle-consistency and adversarial loss functions to facilitate the model training.

\noindent
\textbf{Cycle-consistency Loss.} We apply three types of cycle-consistency loss, \ie, $\mathcal{L}^a_{\mathrm{cyc}}$, $\mathcal{L}^g_{\mathrm{cyc}}$ and $\mathcal{L}^{pix}_{\mathrm{cyc}}$. These three types of cycle-consistency constraints are performed in the geometry space, appearance space and pixel space respectively. Our ablation study in Sec.~\ref{sec:ablationStudy} demonstrates that cycle-consistency constraints are important for pose-preserving in translation.

\noindent
\textbf{Adversarial Loss.} We impose adversarial losses $\mathcal{L}^a_{\mathrm{adv}}$, $\mathcal{L}^g_{\mathrm{adv}}$ and $\mathcal{L}^{pix}_{\mathrm{adv}}$, which correspond to the geometry, appearance and pixel space, respectively. LSGAN is used for more stable training and convergence.

\noindent
\textbf{Total Loss.} In summary, the full loss function of our method is:
\begin{equation}
\vspace{-0.2cm}
\begin{aligned}
\mathcal{L}_{\mathrm{total}} &=
\mathcal{L}_{\mathrm{CVAE}} +
\mathcal{L}_{\mathrm{prior}} +
\mathcal{L}^a_{\mathrm{con}} +
\mathcal{L}^a_{\mathrm{cyc}} \\&+ \mathcal{L}^g_{\mathrm{cyc}} + \mathcal{L}^{pix}_{\mathrm{cyc}} + \mathcal{L}^a_{\mathrm{adv}} +  \mathcal{L}^g_{\mathrm{adv}} +
\mathcal{L}^{pix}_{\mathrm{adv}}
\end{aligned}
\end{equation}
More details of the implementation of these losses are described in the supplementary material.

\begin{table*}[t!]
    \vspace {-0.2cm}
    \caption{\label{table:amt}\small{\textbf{Human perceptual study.} Pairwise A/B tests on horse$\rightarrow$giraffe and human$\rightarrow$cat face task.}}
    \vspace {-0.2cm}
    \begin{subtable}[t]{0.5\textwidth}
        \raggedright
      \resizebox{0.9\columnwidth}{!}{
         \begin{tabular}{c|c|c}
         & \textbf{horse} $\rightarrow$ \textbf{giraffe} & \textbf{human} $\rightarrow$ \textbf{cat face} \\
         Method & \% Testers labeled \textit{better} & \% Testers labeled \textit{better} \\
         \Xhline{1.2pt}
         CycleGAN~\cite{CycleGAN2017} & 15.0$\%$ & 15.4$\%$ \\
         UNIT~\cite{mingyu2017unsupervised} & 19.3$\%$ & 18.9$\%$ \\
         MUNIT~\cite{xun2018multimodal} & 20.4$\%$ & 17.8$\%$ \\
         DRIT~\cite{hsinying2018diverse} & 16.1$\%$ & 23.4$\%$ \\
         Ours & \textbf{50.0$\%$} & \textbf{50.0$\%$} \\
         \end{tabular}
      }
      \caption{\scriptsize{Score of ``realism''.}}
    \end{subtable}%
  \begin{subtable}[t]{0.5\textwidth}
        \raggedleft
        \resizebox{1.0\columnwidth}{!}{
         \begin{tabular}{c|c|c}
         & \textbf{horse} $\rightarrow$ \textbf{giraffe} & \textbf{human} $\rightarrow$ \textbf{cat face} \\
         Method & \% Testers labeled \textit{better} & \% Testers labeled \textit{better} \\
         \Xhline{1.2pt}
         CycleGAN~\cite{CycleGAN2017} & 11.9$\%$ & 25.7$\%$ \\
         UNIT~\cite{mingyu2017unsupervised} & 16.5$\%$ & 23.3$\%$ \\
         MUNIT~\cite{xun2018multimodal} & 19.2$\%$ & 31.7$\%$ \\
         DRIT~\cite{hsinying2018diverse} & 23.6$\%$ &  34.4$\%$ \\
         Ours & \textbf{50.0$\%$} & \textbf{50.0$\%$} \\
         \end{tabular}
      }
        \caption{\scriptsize{Score of ``geometry-consistency''.}}
    \end{subtable}
    \vspace {-0.8cm}
\end{table*}

\section{Experiments}

\noindent
\textbf{Datasets.} We conduct extensive comparisons and ablation studies on four datasets that cover both synthesis and real-world data. (1). Synthesis Animals: We use the publicly available CAD model provided by~\cite{Zuffi_2018_CVPR} to render six different non-rigid animals, \ie, Cheetah, Cow, Lion, Rhino, Bear and Wolf. For each population of animal, we rendered $10,000$ images ($9000$ for training and $1000$ for testing) with different shapes through the randomly sampled parameters. (2). Real-world Animals: We collected $5000$ images ($4500$ for testing and $500$ for testing) of horse and giraffe from Flickr. (3). Unconstrained Face: We collected images of three typical domains, \ie, human, dog and cat faces. We randomly sampled $5,000$ images ($4500$ for testing and $500$ for testing) from YFCC100M~\cite{kalkowski2015real}, Stanford Dog~\cite{aditya2011novel} and CelebA~\cite{ziwei2015deep} datasets respectively. Note that the faces in each dataset are completely unconstrained rather than within four given modes in \cite{xun2018multimodal}.

\noindent
\textbf{Baselines.} We compare our approach with the four most related state-of-the-art methods: CycleGAN~\cite{CycleGAN2017}, UNIT~\cite{mingyu2017unsupervised}, MUNIT~\cite{xun2018multimodal} and DRIT~\cite{hsinying2018diverse}, All of these methods can perform image-to-image translation with unpaired training data. In particular, MUNIT~\cite{xun2018multimodal} and DRIT~\cite{hsinying2018diverse} can generate multimodal results. Thus, we compare to them also in multi-modal generation task. We trained these four baselines on the newly collected datasets with their public implementation with default settings.

\noindent
\textbf{Evaluation Metric.} For quantitative comparison, we evaluate both the realism and diversity of the generated images. Following~\cite{wang2018pix2pixHD,wayne2018reenactgan}, we perform human subjective study for geometry-consistency/realism evaluation. To measure visual quality, rather than general image quality assessment methods~\cite{SSIM,kwanyee2018hallucinated,kwanyee2018self} or perceptual loss~\cite{DBLP:conf/nips/ZhuZPDEWS17}, Fr\'echet Inception Distance (FID)~\cite{anonymous2019large} is adopted. To measure diversity, similar to~\cite{zhu2017toward,xun2018multimodal}, we use the LPIPS metric~\cite{richard2018the} to calculate the distance among images.

\noindent
\textbf{Implementation Details.} Images of all datasets are cropped and resized to $256\times256$. Taking $X$ domain for example. We adopt the architecture for our structure encoder $E^g_x$ from Stack-Hourglass network~\cite{alejandro2016stack} which have shown impressive results for landmark localisation task~\cite{DBLP:conf/cvpr/ChuYOMYW17,DBLP:journals/corr/abs-1712-02765}. For the mapping from $g_x$ to $\hat{x}$ ($E_x^c$ and $D_x$ with skip-connection), we use the UNet architecture~\cite{olaf2015unet} provided by~\cite{CycleGAN2017}. The same architecture of $E_x^c$ is adopted for the appearance encoder $E_x^a$. We use a simple 4-layer fully-connection network followed with ReLU for the transformer $\Psi_{X\leftrightarrow Y}$ and the discriminators. For pixel level adversarial loss, we use the discriminator provided by~\cite{mingyu2017unsupervised}.

We train our model in two main steps. First, to obtain the geometry heatmap $g_x(g_y)$, $E_x^a(E_y^a)$, $E_x^g(E_y^g)$ and $D_x(D_y)$ are trained together for 40 epochs. Then, structure encoders are frozen and all of the networks except $E_x^g$ and $E_y^g$ are trained end-to-end for 20 epochs. We train all of the models use the Adam~\cite{Kingma2014} optimizer with initial $lr = 0.0001$ and $(\beta_1, \beta_2) = (0.5, 0.999)$ on eight NVIDIA V100 GPUs. More details on the training and network architecture are provided in the supplementary material.

\begin{table*}[t!]
\vspace{-0.45cm}
\caption{\small{\textbf{Quantitative Results.} We use FID (lower is better) and diversity (higher is better) with LPIPS distance to evaluate the quality and diversity of the generated images.}}
\label{table:tab_compare_structure}
\vspace{-0.45cm}
\begin{center}
\resizebox{0.9\linewidth}{!}{
\small
\begin{tabular}{c|c|c|c|c|c|c|c|c|c|c|c|c}
& \multicolumn{2}{c|}{\textbf{Real Data}} & \multicolumn{2}{c|}{\textbf{CycleGAN~\cite{CycleGAN2017}}} & \multicolumn{2}{c|}{\textbf{UNIT~\cite{mingyu2017unsupervised}}} & \multicolumn{2}{c|}{\textbf{MUNIT~\cite{xun2018multimodal}}} & \multicolumn{2}{c|}{\textbf{DRIT~\cite{hsinying2018diverse}}} & \multicolumn{2}{c}{\textbf{\textbf{Ours}}}\\
\Xhline{1.2pt}
& FID & Diversity & FID & Diversity & FID & Diversity & FID & Diversity & FID & Diversity & FID & Diversity \\ 
\Xhline{1.2pt}
cats $\rightarrow$ human face  & 0.00 & 0.54 & 57.92 & - &  98.39 & - & 40.91 & \textbf{0.41} & 69.53 & 0.20 & \textbf{32.25} & 0.39 \\
human face $\rightarrow$ cats & 0.00 & 0.65 & 44.23 & - & 35.26 & - & 23.24 & 0.53 & 33.14 & 0.52 & \textbf{21.88} & \textbf{0.56} \\
cats $\rightarrow$ dogs & 0.00 & 0.66 & 143.14 & - & 104.32 & - & 100.26 & 0.59 & 67.01 & 0.54 &\textbf{65.77} & \textbf{0.60} \\
dogs $\rightarrow$ cats & 0.00 & 0.65 & 75.75 & - & 66.84 & - & 27.60 & 0.56 & 31.04 & \textbf{0.59} & \textbf{23.23} & 0.58 \\
dogs $\rightarrow$ human face & 0.00 & 0.54 & 105.09 & - & 103.35 & - & 37.84 & 0.40 & 46.70 & 0.32 & \textbf{31.06} &  \textbf{0.41} \\
human face $\rightarrow$ dogs & 0.00 & 0.66 & 149.61 & - & 91.38 & - & 73.98 & 0.60 & 68.84 & 0.57 & \textbf{52.20} & \textbf{0.67} \\
\Xhline{1.2pt}
Average & 0.00 & 0.62 & 95.96 & - & 83.26 & - & 50.64 & 0.52 & 52.71 & 0.46 & \textbf{37.73}& \textbf{0.54} \\
\end{tabular}
}
\end{center}
\vspace{-0.8cm}
\end{table*}

\subsection{Comparisons with State-of-the-Arts}\label{sec:comparisonState}

\noindent
\textbf{Qualitative Comparison.} Recall the motivation of our work: by introducing the unsupervised latent geometry representation, we hope our framework has a higher capacity for translation between more complicated objects. Here we perform visual quality comparison to state-of-the-art methods in Fig.~\ref{fig:fig_compare_structure}. We evaluate the quality of generated results on both near-rigid (\ie, faces) and non-rigid (animals) objects. Our approach is able to achieve superior results to all of the baselines. Although results of the baselines are recognizable to be settled in the target domain, the geometry tends to be broken due to the neglect of geometry cues. For near-rigid objects, the baselines are likely to yield distorted results. For non-rigid objects, which are more challenging due to the large inter and intra-domain shape variations, the baselines always obtain results with missing parts. By contrast, the translations by our approach are more robust to large shape variations and unconstrained appearance in both rigid and non-rigid scenarios.

For multimodal generation, we compare  our approach with MUNIT~\cite{xun2018multimodal} and DRIT~\cite{hsinying2018diverse} in Fig.~\ref{fig:fig_compare_multimodal}. Both of the baselines can obtain diverse outputs. However, in some unconstrained scenarios, \eg, profile face with sun-glass and large face shape difference between domains, the results of baselines degrade and suffer from severe artifacts. It can be observed that our approach achieve better visual quality than others. More results on other datasets are demonstrated in the supplementary material. 

\noindent
\textbf{Quantitative Comparison.} We use both subjective and objective metrics for the quantitative performance evaluation. For the realism of generation images, we ask volunteers to perform subjective pairwise A/B tests. 
Following the metric in MUNIT~\cite{xun2018multimodal}, the preference score of our work indicates the percentage that one method (CycleGAN~\cite{CycleGAN2017}, UNIT~\cite{mingyu2017unsupervised}, MUNIT~\cite{xun2018multimodal}, DRIT~\cite{hsinying2018diverse}) is preferred over \textit{our method}. For each time of a test, participants can vote for A/B/Not Sure. Two metrics are evaluated as shown in Table.~\ref{table:amt}, \textit{realism} for evaluation the similarity with real data while \textit{geometry-consistency} for evaluation the geometry consistency with input image. Participants were given 10 seconds to choose which image has better \textit{realism} or \textit{geometry-consistency} in a pair of generated images from two different methods. All 500 test images of each dataset are compared 100 times by different participants. Our method obtains the highest preference rate.

For the evaluation of visual quality and diversity, following~\cite{junyan2017toward}, we use 100 input images in the test set and sample 19 output pairs per input. We compute the average LPIPS distance in ImageNet pre-trained AlexNet feature space between the $1,900$ pairs of images. FID is calculated between the real data and the generated results. As shown in Table~\ref{table:tab_compare_structure}, our method significantly outperforms all of the baselines both in visual quality and diversity. In particular, even though MUNIT and DRIT obtain reasonable performance in diversity, they get a poor score in the subjective metric, suggesting the shortcoming of these methods in handling translation across a large geometry gap.

\begin{figure}[t!]
\centering
\includegraphics[width=0.9\linewidth]{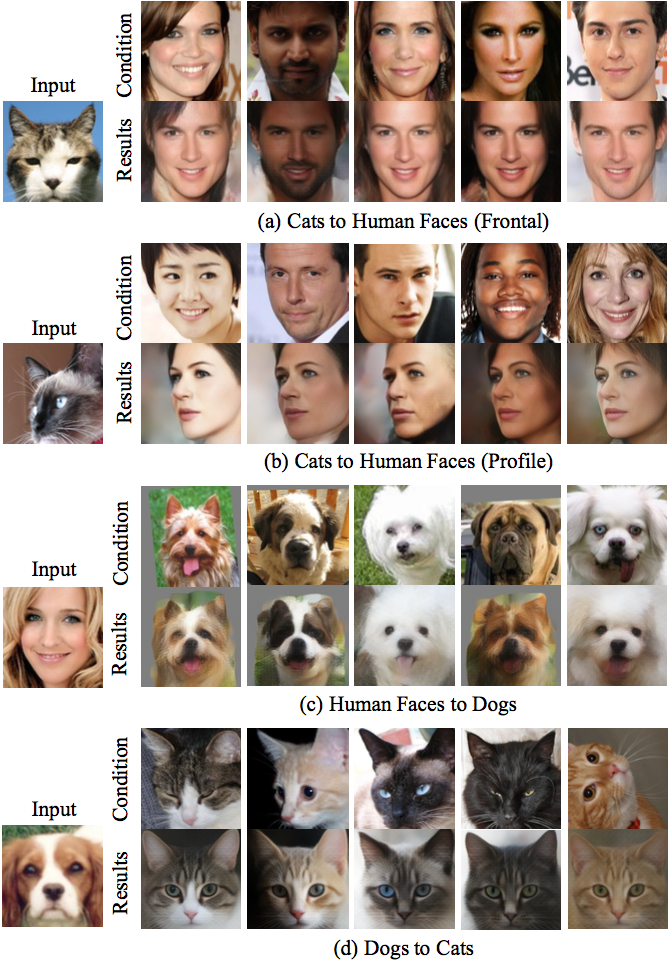}
\vspace{-0.5cm}
\caption{\small{\textbf{Exemplar-guided generation.} Conditional generation with different images as appearance reference on cat$\rightarrow$human face, human$\rightarrow$dog face, dog$\rightarrow$cat face tasks.}}
\vspace{-0.5cm}
\label{fig:fig_exp_guided}
\end{figure}

\begin{figure*}[t]
\centering
\includegraphics[width=0.95\linewidth]{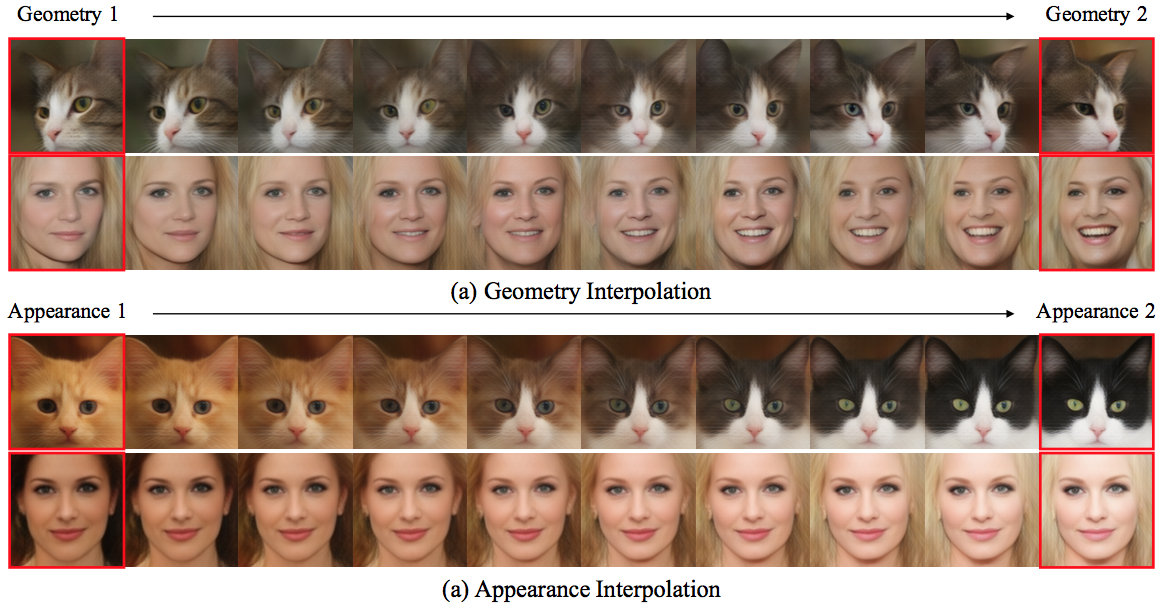}
\vspace{-0.35cm}
\caption{\small{\textbf{Interpolation.} Linear interpolation results of geometry and appearance latent code on cat and human face datasets.}}
\label{fig:fig_interpolation}
\vspace{-0.35cm}
\end{figure*}

\begin{figure}[h]
\centering
\includegraphics[width=0.95\linewidth]{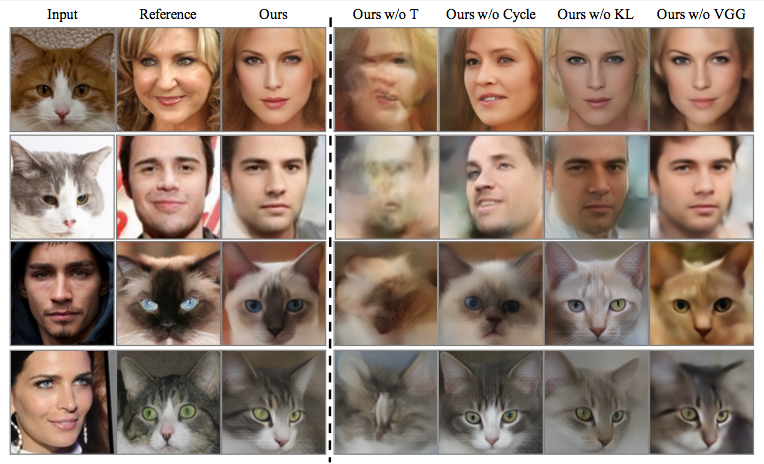}
\vspace{-0.25cm}
\caption{\small{\textbf{Quantitative ablation study.} Visualisation results on human$\leftrightarrow$cat task.}}
\label{fig:fig_ablation}
\vspace{-0.2cm}
\end{figure}

\subsection{Representation Disentanglement}\label{sec:disentagleExperiment}
\noindent
\textbf{Exemplar-guided Image Translation.} In Fig.~\ref{fig:fig_exp_guided}, we illustrate the exemplar-guided translation results of several typical shapes of faces, \eg, frontal, profile, eye-closed and mouth-opening. From input to output, we observe that the geometry feature maintains faithfully. Thanks to the pure geometry representation translation schema, which endow the model with the ability of appearance-agnostic image-to-image translation. In addition, once the geometry is translated successfully, the model can take images in the target domain as exemplars to guide multimodal generations. Results in Fig.~\ref{fig:fig_exp_guided} show successful \textit{disentanglement} of the geometry and appearance in two aspects. First, the geometry maintains to be the same no matter what shape of the exemplar is. As a concrete example, as shown in Fig.~\ref{fig:fig_exp_guided} (b), the generated faces maintain to be profile even with large variations of the exemplars. Second, the appearance of exemplars can be successfully transferred to the generated images, even for the detail textures, \eg, the beard of the man in Fig.~\ref{fig:fig_exp_guided} (a) and the blue eyes of the cat in Fig.~\ref{fig:fig_exp_guided} (d).

\noindent
\textbf{Interpolation.} To evaluate whether disentangled latent space is densely populated, we perform a linear interpolation to geometry code and appearance code respectively in Fig.~\ref{fig:fig_interpolation}. The interpolation results show that both the geometry and appearance of images can change smoothly along with the latent space from source to target. It is noteworthy that the datasets have only one geometry and appearance for each sample and only discrete features are supplied from standalone individuals in raw datasets. The smooth interpolation results show that our model has captured a reasonable coverage of the manifold successfully.

\begin{table}[h]
\vspace {-0.2cm}
\caption{\label{table:ablation} \small{\textbf{Ablation study.} Fooling rate of ``real vs fake''.}}
\vspace {-0.5cm}
\begin{center}
\resizebox{1.0\columnwidth}{!}{
\begin{tabular}{c|c|c}
& \textbf{horse} $\rightarrow$ \textbf{giraffe} & \textbf{human} $\rightarrow$ \textbf{cat face} \\
Method & \% Testers labeled \textit{real} & \% Testers labeled \textit{real} \\
\Xhline{1.2pt}
Ours w/o Trans. & 0.0$\%$ & 0.0$\%$ \\
Ours w/o $\mathcal{L}_{\text{cyc}}$ & 14.2$\%$ & 16.8$\%$ \\
Ours w/o KL & 10.2$\%$ & 15.4$\%$ \\
Ours w/o VGG & 12.6$\%$ & 18.4$\%$ \\
Ours & \textbf{16.2}$\%$ & \textbf{23.9}$\%$ \\
\end{tabular}
}
\end{center}
\vspace {-0.8cm}
\end{table}

\subsection{Ablation Study}\label{sec:ablationStudy}
To isolate the effectiveness of pivotal components of our method, we perform an ablation study on the quality of generated images. We evaluate several variants of our method: 1) Ours w/o T: our methods without appearance and geometry transformers. 2) Ours w/o cycle: our methods without the cycle-consistency loss term. 3) Ours w/o KL: our methods without the KL loss term. 4) Ours w/o VGG: replacing VGG loss with L1 loss in our method.

Figure~\ref{fig:fig_ablation} shows the qualitative results of the variants. Without the transformer, our method is unable to generate plausible results to cross the large gap of the geometry representation between two domains. Without using the cycle-consistency loss, our method can still obtain plausible results. However, the pose-consistency with the input image cannot be guaranteed, suggesting that the cycle-consistency loss is a key component for pose-preserving. 
Without using the KL loss, the consistency with the reference image cannot be maintained. Without the VGG loss, we obtain blurred results, which is consistent with the observation of~\cite{DBLP:conf/iccv/SajjadiSH17,esser2018variational}.

We quantify these observations with perceptual studies on the human$\rightarrow$cat face and horse$\rightarrow$giraffe task in Table~\ref{table:ablation}. The scores obtained by our method on these two tasks demonstrate its capability in generating realistic results. Note that without cycle-consistency loss, a comparable perceptual score can also be achieved with our method, which indicates that this loss is more important for pose-preserving than generated quality.

\vspace{-0.3cm}

\section{Conclusion}
We have presented a novel geometry-aware disentangle-and-translate framework for image-to-image translation, in which we introduced an unsupervised geometry latent branch based on CycleGAN system. Specifically, we first disentangled each domain on the geometry space and appearance space, then established the translation on each latent space. Our extensive qualitative and quantitative experiments showed that our method is effective for translation between objects with complex structures. Moreover, our model can also support multimodal translation and outperform previous state-of-the-art methods. Future work includes extending this framework to more unconstrained scenarios, such as images in ImageNet and YouTube videos.

\noindent
\textbf{Acknowledgement.} We would like to thank Kwan-Yee Lin and Jingtan Piao for insightful discussion and their exceptional support.

{\small
\bibliographystyle{ieee_fullname}
\bibliography{TransGaGa}

\begin{thebibliography}{10}\itemsep=-1pt

\bibitem{Balakrishnan_2018_CVPR}
Guha Balakrishnan, Amy Zhao, Adrian~V. Dalca, Fr{\'e}do Durand, and John
  Guttag.
\newblock Synthesizing images of humans in unseen poses.
\newblock In {\em CVPR}, 2018.

\bibitem{anonymous2019large}
Andrew Brock, Jeff Donahue, and Karen Simonyan.
\newblock Large scale gan training for high fidelity natural image synthesis.
\newblock In {\em ICLR}, 2019.

\bibitem{DBLP:journals/corr/abs-1712-02765}
Adrian Bulat and Georgios Tzimiropoulos.
\newblock Super-fan: Integrated facial landmark localization and
  super-resolution of real-world low resolution faces in arbitrary poses with
  gans.
\newblock In {\em CVPR}, 2018.

\bibitem{cao2018cari}
Kaidi Cao, Jing Liao, and Lu Yuan.
\newblock Carigans: Unpaired photo-to-caricature translation.
\newblock In {\em Siggraph Asia}, 2018.

\bibitem{chen2016infogan}
Xi Chen, Yan Duan, Rein Houthooft, John Schulman, Ilya Sutskever, and Pieter
  Abbeel.
\newblock Infogan: Interpretable representation learning by information
  maximizing generative adversarial nets.
\newblock In {\em NIPS}, 2016.

\bibitem{chen2019self}
Xipeng Chen, Kwan-Yee Lin, Wentao Liu, Chen Qian, and Liang Lin.
\newblock Weakly-supervised discovery of geometry-aware representation for 3d
  human pose estimation.
\newblock In {\em CVPR}, 2019.

\bibitem{DBLP:conf/cvpr/ChuYOMYW17}
Xiao Chu, Wei Yang, Wanli Ouyang, Cheng Ma, Alan~L. Yuille, and Xiaogang Wang.
\newblock Multi-context attention for human pose estimation.
\newblock In {\em CVPR}, 2017.

\bibitem{esser2018variational}
Patrick Esser, Ekaterina Sutter, and Bj{\"o}rn Ommer.
\newblock A variational u-net for conditional appearance and shape generation.
\newblock In {\em CVPR}, 2018.

\bibitem{DBLP:conf/nips/GatysEB15}
Leon~A. Gatys, Alexander~S. Ecker, and Matthias Bethge.
\newblock Texture synthesis using convolutional neural networks.
\newblock In {\em NIPS}, 2015.

\bibitem{gokaslan2018improving}
Aaron Gokaslan, Vivek Ramanujan, Daniel Ritchie, Kwang~In Kim, and James
  Tompkin.
\newblock Improving shape deformation in unsupervised image-to-image
  translation.
\newblock {\em ECCV}, 2018.

\bibitem{ian2014generative}
Ian~J. Goodfellow, Jean Pouget{-}Abadie, Mehdi Mirza, Bing Xu, David
  Warde{-}Farley, Sherjil Ozair, Aaron~C. Courville, and Yoshua Bengio.
\newblock Generative adversarial nets.
\newblock In {\em NIPS}, 2014.

\bibitem{DBLP:conf/cvpr/HeZRS16}
Kaiming He, Xiangyu Zhang, Shaoqing Ren, and Jian Sun.
\newblock Deep residual learning for image recognition.
\newblock In {\em CVPR}, 2016.

\bibitem{hadditioniggins2017beta}
Irina Higgins, Loic Matthey, Arka Pal, Christopher Burgess, Xavier Glorot,
  Matthew Botvinick, Shakir Mohamed, and Alexander Lerchner.
\newblock beta-vae: Learning basic visual concepts with a constrained
  variational framework.
\newblock In {\em ICLR}, 2017.

\bibitem{xun2018multimodal}
Xun Huang, Ming{-}Yu Liu, Serge~J. Belongie, and Jan Kautz.
\newblock Multimodal unsupervised image-to-image translation.
\newblock In {\em ECCV}, 2018.

\bibitem{isola2017image}
Phillip Isola, Jun-Yan Zhu, Tinghui Zhou, and Alexei~A Efros.
\newblock Image-to-image translation with conditional adversarial networks.
\newblock In {\em CVPR}, 2017.

\bibitem{jakab2018conditional}
Tomas Jakab, Ankush Gupta, Hakan Bilen, and Andrea Vedaldi.
\newblock Conditional image generation for learning the structure of visual
  objects.
\newblock In {\em NeurIPS}, 2018.

\bibitem{DBLP:conf/eccv/JohnsonAF16}
Justin Johnson, Alexandre Alahi, and Li Fei{-}Fei.
\newblock Perceptual losses for real-time style transfer and super-resolution.
\newblock In {\em ECCV}, 2016.

\bibitem{kalkowski2015real}
Sebastian Kalkowski, Christian Schulze, Andreas Dengel, and Damian Borth.
\newblock Real-time analysis and visualization of the yfcc100m dataset.
\newblock In {\em MM Workshop}, 2015.

\bibitem{aditya2011novel}
Aditya Khosla, Nityananda Jayadevaprakash, Bangpeng Yao, and Li Fei-Fei.
\newblock Novel dataset for fine-grained image categorization.
\newblock In {\em CVPR Workshop}, 2011.

\bibitem{taeksoo2017learning}
Taeksoo Kim, Moonsu Cha, Hyunsoo Kim, Jung~Kwon Lee, and Jiwon Kim.
\newblock Learning to discover cross-domain relations with generative
  adversarial networks.
\newblock In {\em ICML}, 2017.

\bibitem{Kingma2014}
Diederik~P. Kingma and Max Welling.
\newblock Auto-encoding variational bayes.
\newblock In {\em ICLR}, 2014.

\bibitem{DBLP:conf/cvpr/LedigTHCCAATTWS17}
Christian Ledig, Lucas Theis, Ferenc Huszar, Jose Caballero, Andrew Cunningham,
  Alejandro Acosta, Andrew~P. Aitken, Alykhan Tejani, Johannes Totz, Zehan
  Wang, and Wenzhe Shi.
\newblock Photo-realistic single image super-resolution using a generative
  adversarial network.
\newblock In {\em CVPR}, 2017.

\bibitem{hsinying2018diverse}
Hsin{-}Ying Lee, Hung{-}Yu Tseng, Jia{-}Bin Huang, Maneesh Singh, and
  Ming{-}Hsuan Yang.
\newblock Diverse image-to-image translation via disentangled representations.
\newblock In {\em ECCV}, 2018.

\bibitem{Lin_2018_CVPR}
Jianxin Lin, Yingce Xia, Tao Qin, Zhibo Chen, and Tie-Yan Liu.
\newblock Conditional image-to-image translation.
\newblock In {\em CVPR}, 2018.

\bibitem{kwanyee2018hallucinated}
Kwan-Yee Lin and Guangxiang Wang.
\newblock Hallucinated-iqa: No-reference image quality assessment via
  adversarial learning.
\newblock In {\em CVPR}, 2018.

\bibitem{kwanyee2018self}
Kwan-Yee Lin and Guangxiang Wang.
\newblock Self-supervised deep multiple choice learning network for blind image
  quality assessment.
\newblock In {\em BMVC}, 2018.

\bibitem{mingyu2017unsupervised}
Ming{-}Yu Liu, Thomas Breuel, and Jan Kautz.
\newblock Unsupervised image-to-image translation networks.
\newblock In {\em NIPS}, 2017.

\bibitem{ziwei2015deep}
Ziwei Liu, Ping Luo, Xiaogang Wang, and Xiaoou Tang.
\newblock Deep learning face attributes in the wild.
\newblock In {\em ICCV}, 2015.

\bibitem{liqian2018exemplar}
Liqian Ma, Xu Jia, Stamatios Georgoulis, Tinne Tuytelaars, and Luc~Van Gool.
\newblock Exemplar guided unsupervised image-to-image translation.
\newblock In {\em NeurIPS}, 2018.

\bibitem{ma2018disentangled}
Liqian Ma, Qianru Sun, Stamatios Georgoulis, Luc Van~Gool, Bernt Schiele, and
  Mario Fritz.
\newblock Disentangled person image generation.
\newblock In {\em CVPR}, 2018.

\bibitem{alejandro2016stack}
Alejandro Newell, Kaiyu Yang, and Jia Deng.
\newblock Stacked hourglass networks for human pose estimation.
\newblock In {\em ECCV}, 2016.

\bibitem{oord2016pixel}
Aaron van~den Oord, Nal Kalchbrenner, and Koray Kavukcuoglu.
\newblock Pixel recurrent neural networks.
\newblock 2016.

\bibitem{olaf2015unet}
Olaf Ronneberger, Philipp Fischer, and Thomas Brox.
\newblock U-net: Convolutional networks for biomedical image segmentation.
\newblock In {\em MICCAI}, 2015.

\bibitem{DBLP:conf/iccv/SajjadiSH17}
Mehdi S.~M. Sajjadi, Bernhard Sch{\"{o}}lkopf, and Michael Hirsch.
\newblock Enhancenet: Single image super-resolution through automated texture
  synthesis.
\newblock In {\em ICCV}, 2017.

\bibitem{DBLP:journals/corr/SimonyanZ14a}
Karen Simonyan and Andrew Zisserman.
\newblock Very deep convolutional networks for large-scale image recognition.
\newblock {\em CoRR}, abs/1409.1556, 2014.

\bibitem{sohn2015learning}
Kihyuk Sohn, Honglak Lee, and Xinchen Yan.
\newblock Learning structured output representation using deep conditional
  generative models.
\newblock In {\em NIPS}, 2015.

\bibitem{DBLP:conf/nips/ThewlisBV17}
James Thewlis, Hakan Bilen, and Andrea Vedaldi.
\newblock Unsupervised learning of object frames by dense equivariant image
  labelling.
\newblock In {\em NIPS}, 2017.

\bibitem{james2017unsupervised}
James Thewlis, Hakan Bilen, and Andrea Vedaldi.
\newblock Unsupervised learning of object landmarks by factorized spatial
  embeddings.
\newblock In {\em ICCV}, 2017.

\bibitem{NIPS2016_6527}
Aaron van~den Oord, Nal Kalchbrenner, Lasse Espeholt, koray kavukcuoglu, Oriol
  Vinyals, and Alex Graves.
\newblock Conditional image generation with pixelcnn decoders.
\newblock In {\em NIPS}, 2016.

\bibitem{wang2018vid2vid}
Ting-Chun Wang, Ming-Yu Liu, Jun-Yan Zhu, Guilin Liu, Andrew Tao, Jan Kautz,
  and Bryan Catanzaro.
\newblock Video-to-video synthesis.
\newblock In {\em NeurIPS}, 2018.

\bibitem{wang2018pix2pixHD}
Ting-Chun Wang, Ming-Yu Liu, Jun-Yan Zhu, Andrew Tao, Jan Kautz, and Bryan
  Catanzaro.
\newblock High-resolution image synthesis and semantic manipulation with
  conditional gans.
\newblock In {\em CVPR}, 2018.

\bibitem{Wang_2018_CVPR}
Wei Wang, Xavier Alameda-Pineda, Dan Xu, Pascal Fua, Elisa Ricci, and Nicu
  Sebe.
\newblock Every smile is unique: Landmark-guided diverse smile generation.
\newblock In {\em CVPR}, 2018.

\bibitem{wang2018esrgan}
Xintao Wang, Ke Yu, Shixiang Wu, Jinjin Gu, Yihao Liu, Chao Dong, Yu Qiao, and
  Chen~Change Loy.
\newblock {ESRGAN:} enhanced super-resolution generative adversarial networks.
\newblock In {\em ECCV Workshop}, 2018.

\bibitem{SSIM}
Zhou Wang, Alan~C. Bovik, Hamid~R. Sheikh, and Eero~P. Simoncelli.
\newblock Image quality assessment: from error visibility to structural
  similarity.
\newblock {\em IEEE Transactions on Image Processing (TIP)}, 13(4):600--612,
  2004.

\bibitem{wayne2018reenactgan}
Wayne Wu, Yunxuan Zhang, Cheng Li, Chen Qian, and Chen~Change Loy.
\newblock Reenactgan: Learning to reenact faces via boundary transfer.
\newblock In {\em ECCV}, 2018.

\bibitem{zili2017dualgan}
Zili Yi, Hao~(Richard) Zhang, Ping Tan, and Minglun Gong.
\newblock Dualgan: Unsupervised dual learning for image-to-image translation.
\newblock In {\em ICCV}, 2017.

\bibitem{DBLP:conf/eccv/ZhangIE16}
Richard Zhang, Phillip Isola, and Alexei~A. Efros.
\newblock Colorful image colorization.
\newblock In {\em ECCV}, 2016.

\bibitem{richard2018the}
Richard Zhang, Phillip Isola, Alexei~A. Efros, Eli Shechtman, and Oliver Wang.
\newblock The unreasonable effectiveness of deep features as a perceptual
  metric.
\newblock In {\em CVPR}, 2018.

\bibitem{Zhang_2018_CVPR}
Yuting Zhang, Yijie Guo, Yixin Jin, Yijun Luo, Zhiyuan He, and Honglak Lee.
\newblock Unsupervised discovery of object landmarks as structural
  representations.
\newblock In {\em CVPR}, 2018.

\bibitem{DBLP:conf/nips/ZhuZPDEWS17}
Jun{-}Yan Zhu, Richard Zhang, Deepak Pathak, Trevor Darrell, Alexei~A. Efros,
  Oliver Wang, and Eli Shechtman.
\newblock Toward multimodal image-to-image translation.
\newblock In {\em NIPS}, 2017.

\bibitem{junyan2017toward}
Jun{-}Yan Zhu, Richard Zhang, Deepak Pathak, Trevor Darrell, Alexei~A. Efros,
  Oliver Wang, and Eli Shechtman.
\newblock Toward multimodal image-to-image translation.
\newblock In {\em NIPS}, 2017.

\bibitem{CycleGAN2017}
Jun-Yan Zhu, Taesung Park, Phillip Isola, and Alexei~A Efros.
\newblock Unpaired image-to-image translation using cycle-consistent
  adversarial networkss.
\newblock In {\em ICCV}, 2017.

\bibitem{zhu2017toward}
Jun-Yan Zhu, Richard Zhang, Deepak Pathak, Trevor Darrell, Alexei~A Efros,
  Oliver Wang, and Eli Shechtman.
\newblock Toward multimodal image-to-image translation.
\newblock In {\em NIPS}, 2017.

\bibitem{Zuffi_2018_CVPR}
Silvia Zuffi, Angjoo Kanazawa, and Michael~J. Black.
\newblock Lions and tigers and bears: Capturing non-rigid, 3d, articulated
  shape from images.
\newblock In {\em CVPR}, 2018.

\end{thebibliography}
}

\end{document}